\documentclass{article}
\usepackage{spconf,amsmath,epsfig}

\usepackage{graphicx}
\usepackage{todonotes}
\usepackage{amsfonts}

\usepackage{graphicx} 
\usepackage{epsfig} 

\usepackage{booktabs}
\usepackage{siunitx}
\usepackage{soul,framed}
\usepackage{color}
\usepackage[normalem]{ulem} 

\usepackage{colortbl}
\usepackage{arydshln}
\usepackage[flushleft]{threeparttable}
\usepackage{algorithm,algorithmic}

\definecolor{darkgreen}{rgb}{0,0.6,0.2}

\title{Self-Constructing Graph Convolutional Networks \\for Semantic Labeling}
\name{Author(s) Name(s)\thanks{Thanks to XYZ agency for funding.}}
\address{Author Affiliation(s)}

\name{Qinghui Liu$^{1,2}$, Michael Kampffmeyer$^{2}$, Robert Jenssen$^{2,1}$, Arnt-B{\o}rre Salberg$^{1}$ \thanks{This  work  is  supported  by  the  foundation  of  the  Research Council of Norway under Grant 220832.}}
\address{$^{1}$Norwegian Computing Center, Dept. SAMBA, NO-0314 OSLO, Norway \\
$^{2}$UiT Machine Learning Group, UiT the Arctic University of Norway, Troms{\o}, Norway}%

\begin{document}
%
\maketitle
\begin{abstract}
Graph Neural Networks (GNNs) have received increasing attention in many fields. However, due to the lack of prior graphs, their use for semantic labeling has been limited. Here, we propose a novel architecture called the Self-Constructing Graph (SCG), which makes use of learnable latent variables to generate embeddings and to self-construct the underlying graphs directly from the input features without relying on manually built prior knowledge graphs. SCG can automatically obtain optimized non-local context graphs from complex-shaped objects in aerial imagery. We optimize SCG via an adaptive diagonal enhancement method and a variational lower bound that consists of a customized graph reconstruction term and a Kullback-Leibler divergence regularization term. We demonstrate the effectiveness and flexibility of the proposed SCG on the publicly available ISPRS Vaihingen dataset and our model SCG-Net achieves competitive results in terms of F1-score with much fewer parameters and at a lower computational cost compared to related pure-CNN based work.
\end{abstract}
\section{Introduction}

Recently, graph neural networks (GNNs) \cite{ bronstein2017geometric} and Graph Convolutional Networks (GCNs) \cite{kipf2016semi} 
have received increasing attention, partially due to their superior performance for many node or graph classification tasks in the non-Euclidean domain, including graphs and manifolds. 
Variants of GNNs and GCNs have been
applied to computer vision tasks, among others, image classification \cite{knyazev2019image},  few-shot and zero-shot classification \cite{kampffmeyer2019rethinking}, point clouds classification \cite{wang2019dynamic} and semantic segmentation \cite{liang2018symbolic}. 
However, graph reasoning for vision tasks is quite sensitive to how the graph of relations between objects is built and previous approaches commonly rely on manually built graphs based on prior knowledge.
Inspired by variational graph auto-encoders \cite{kipf2016variational}, we instead propose a novel Self-Constructing Graph module (SCG) to \emph{learn} how a 2D feature map can be transformed into a latent graph structure and how pixels can be assigned to the vertices of the graph from the available training data. In our proposed self-constructing graph convolutional network (SCG-Net), the SCG is followed by Graph Convolutional Networks (GCNs) \cite{kipf2016semi} to update the node features along the edges of the graph. After K-layer of GCNs, the vertices are projected back onto the 2D plane. The SCG module can be easily embedded into existing CNN and GCN networks for computer vision tasks. Our model can be trained end-to-end since every step is fully differentiable. Our experiments demonstrate that the network achieves robust and competitive results on the representative ISPRS 2D semantic labeling Vaihingen benchmark datasets \cite{ISPRS2018}.

\section{Methods}
\label{method}
We first briefly revisit some concepts of graph convolutions. We then present the details of the proposed self-constructing graph (SCG) algorithm and our end-to-end trainable model SCG-Net for semantic labeling tasks.

\begin{figure*}[htbp]
 \centering
  \includegraphics[width=0.70\textwidth]{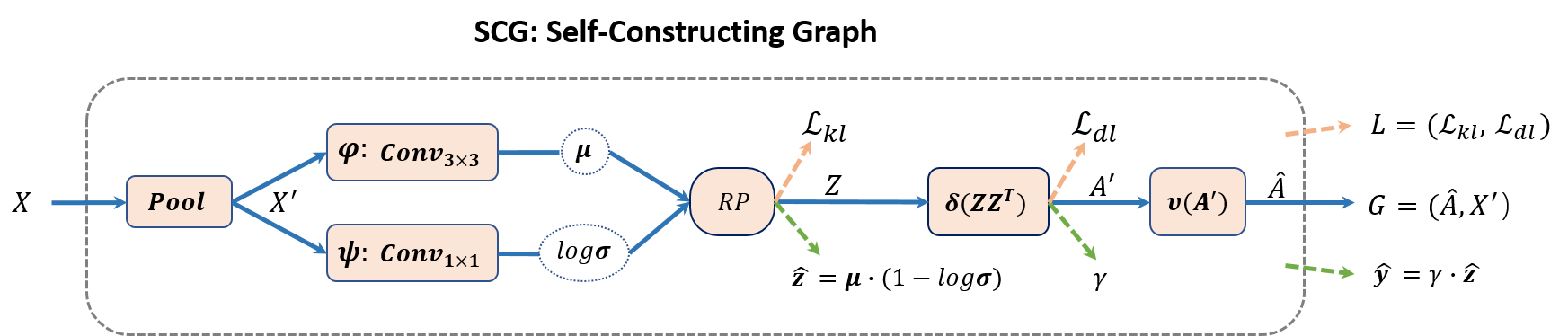} 
  \caption{The illustration diagram of the SCG module, where $\hat{A}$ is the normalized adjacency matrix, $X^\prime$ corresponds to node features, $Z$ are the latent embeddings, $\varepsilon$ are learnable weights, $\mathcal{L}_{kl}$ is the Kullback–Leibler divergence loss and $\mathcal{L}_{dl}$ is the diagonal log loss. $\gamma$ is the adaptive factor, $\hat{\boldsymbol{z}}$ are auxiliary embeddings, $RP$ is the re-parameterization operations and $\hat{\boldsymbol{y}}$ are residual predictions.}
  \label{fig:scg_module}
\end{figure*}

\subsection{Graph Convolution}
\textbf{Definitions}:
We consider an undirected graph $G = (A, X)$, which consists of $n$ vertices, where $A \in \mathbb{R}^{n \times n}$ is the adjacency matrix with $(i,j)$ entry $A_{ij}$ is $1$ if there is an edge between $i$ and $j$ and $0$ otherwise, and $X \in \mathbb{R}^{n \times d}$ is the node feature matrix for all vertices assuming each node has $d$ features. Given a set of labeled nodes $D = \{(x_i, y_i)\}_{i=1}^{m}$ where $x_i \in X$, and $y_i \in Y$ contains the labels for the labeled nodes. 

\textbf{GCN}:
Graph Convolutional Networks (GCNs) \cite{kipf2016semi}  were originally proposed for semi-supervised classification ($m = \left|{Y}\right| \leq n$). Thus, $m = \left|{Y}\right| \leq n$ for semi-supervised node classification settings. GCN implements a "message-passing" function by a combination of linear transformations over one-hop neighbourhoods followed by a non-linearity: 

\begin{equation} \label{eq:gcn2}
Z^{(l+1)}=\sigma\left(\hat{A} X^{(l)} \theta^{(l)}\right) \; ,
\end{equation}

where $\hat{A}$ is the symmetric normalization of $A$ with self-loops:

\begin{equation} \label{eq:norm}
\hat{A}=D^{-\frac{1}{2}}(A+I)D^{\frac{1}{2}} \; ,
\end{equation}

where $D_{ii} = \sum_{j}(A+I)_{ij}$ is a diagonal matrix with node degrees and $I$ is an identity matrix, and $\sigma$ denotes the non-linearity function (e.g. $ReLU$). 

In the following sections, we will use $Z^{(K)} = \operatorname{GCN}(A, X)$ to denote an arbitrary $\operatorname{GCN}$ module implementing $K$ steps of message passing based on some adjacency matrix $A$ and input node features $X$, where $K$ is common in the range $2\text{-}6$ in practice. 

\subsection{Self-Constructing Graph}
We propose the Self-Constructing Graph (SCG) framework for learning latent graph representations directly from 2D feature maps. This model makes use of re-parameterized latent variables and is capable of constructing undirected graphs without relying on prior graph information (see Figure~\ref{fig:scg_module}). We assume an input 2D feature map $X$ of size $h \times w$ with $d$ features. $X \in \mathbb{R}^{h \times w \times d}$ are usually the high-level features learned by deep convolutional networks. The main goal of our SCG module is to learn a latent graph from the input feature maps to capture the long-range relations among vertices. Formally, $G = \operatorname{SCG}(X) $, where $G = (\hat{A}, X^{\prime})$, and $\hat{A} \in \mathbb{R}^{n \times n}$ is a weighted adjacency matrix, $X^{\prime} \in \mathbb{R}^{n \times d}$ is the node features, and  $n = h^{\prime} \times w^{\prime}$ denotes the number of nodes.  Note that usually $(h^\prime \times w^\prime) \leq (h \times w)$ in practice. 


In this work, we take a parameter-free pooling operation (e.g. adaptive avg pooling) to transform $X$ to $X^\prime$ and then constraint the size of vertices to be $n$.

\textbf{Encoder to a latent space}: In the encoding part, Gaussian parameters (the mean matrix $\boldsymbol{\mu} \in \mathbb{R}^{n \times c}$ and the standard deviation matrix $\boldsymbol{\sigma} \in \mathbb{R}^{n \times c}$, where $c$ denotes the number of labels.) are learned from two single-layer $\operatorname{Conv}$ networks (where the subscript indicates of $\operatorname{Conv}$ denote the size of the filters as shown in the following two formulas). 
$$\boldsymbol{\mu} \leftarrow \varphi(X^{\prime}) = \operatorname{Conv}_{3 \times 3}(X^\prime)$$
$$\boldsymbol{\sigma} \leftarrow \exp \left(\psi\left(X^{\prime}\right)\right)= \exp \left(\operatorname{Conv}_{1 \times 1}(X^\prime)\right)$$

Note that $X^\prime$ is compatible with regular convolutional networks by simply reshaping it from $\mathbb{R}^{n \times d}$ to $\mathbb{R}^{h^{\prime} \times w^{\prime} \times d}$. And similarly, the outputs of $\operatorname{Conv}$ are reshaped back from $\mathbb{R}^{h^{\prime} \times w^{\prime} \times c}$ to $\mathbb{R}^{n \times c}$.

\textbf{Reparameterization}: In order to keep the proposed architecture end-to-end trainable, we perform a reparameterization of the latent embeddings. The latent embedding $Z$ is computed as 
$$\textit{Z} \leftarrow \boldsymbol{\mu} + \boldsymbol{\sigma} \cdot  \boldsymbol{\varepsilon}$$ \; ,
where $\varepsilon \in \mathbb{R}^{N^{\prime} \times C}$ is an auxiliary noise variable that is initialized from a standard normal distribution ($\boldsymbol{\varepsilon} \sim \operatorname{N}(0,I)$).

Here, we also introduce auxiliary embeddings $\boldsymbol{\hat{z}}$ which is defined as: $\boldsymbol{\hat{z}} \leftarrow \boldsymbol{\mu} \cdot (1 - {\log \boldsymbol{\sigma}})$, which will be used later to computer the residual predictions. 

We further, during the training phase, regularize the latent variables by minimizing the Kullback-Leibler divergence between the embedding and a centered isotropic multivariate Gaussian prior distribution
\cite{kingma2013auto} $\mathcal{L}_{kl} \leftarrow KL(\boldsymbol{\mu}, \boldsymbol{\sigma})$, which is given as

\begin{equation}\label{eq:klloss}
KL(\boldsymbol{\mu}, \boldsymbol{\sigma}) \simeq -\frac{1}{2n} \sum_{i=1}^{n}\left(1+\log \left(\sigma_{i} \right)^{2}-\mu_{i}^{2}-\sigma_{i}^{2}\right) \; .
\end{equation}

\textbf{Decoder to output space}:
The learned graph $A^\prime$ is given by the inner product between the latent embeddings
$$A^{\prime} \leftarrow \delta(Z Z^{T}) = \operatorname{ReLU}(Z Z^{T})$$ \; .

Note, $A_{ij}^{\prime} > 0$ denotes that there is an edge between nodes $i$ and $j$. Intuitively, we consider $A_{ii}^{\prime}$ shall be $>0$. We therefore introduce the following diagonal log regularization term
\begin{equation}\label{eq:dlloss}
  \mathcal{L}_{dl} \leftarrow  DL(A^\prime) =- \frac{\gamma}{n^2} \sum_{i=1}^{n} \operatorname{log} (\left|A_{ii}^\prime\right|_{[0,1]} + \epsilon)  \; ,
\end{equation}
where $\gamma$ is an adaptive factor which is defined as
\begin{equation}\label{eq:lambda}
    \gamma = \sqrt{1 + \frac{n}{ \sum_{i=1}^{n} (A_{ii}^\prime) + \epsilon}} \; .
\end{equation}

We also propose an adaptive diagonal enhancement approach to better maintain the learned neighborhoods information resulting in
\begin{equation}\label{eq:enhance}
    A^\prime \gets A^\prime + \gamma \cdot \operatorname{diag}(A^\prime) \; .
\end{equation}

We finally obtain the symmetric normalized $\hat{A}$ w.r.t the enhanced $A^\prime$ by
\begin{equation}\label{eq:normenhance}
\hat{A} \leftarrow \nu(A^{\prime}) = D^{-\frac{1}{2}}\left(A^\prime+\gamma \cdot \operatorname{diag}\left(A^\prime\right)+I\right)D^{\frac{1}{2}} \; .
\end{equation}

Additionally, we also propose a residual term, the so-called adaptive residual prediction $\boldsymbol{\hat{y}}$ which is defined as: $\boldsymbol{\hat{y}} \leftarrow \gamma \cdot \boldsymbol{\hat{z}}$, to be used later for refining the final predictions of the networks. 

\subsection{The SCG-Net} \label{archit}

\begin{figure}[thpb!]
 \centering
  \includegraphics[width=0.99\columnwidth]{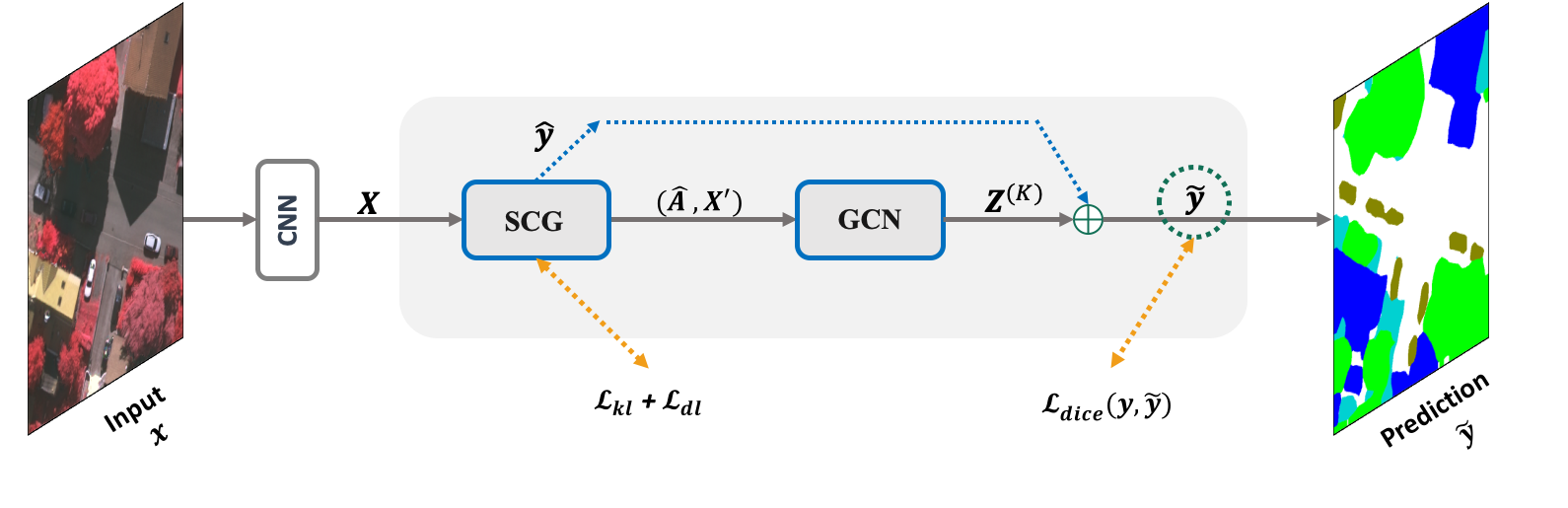} 
  \caption{Model architecture of SCG-Net for semantic labeling includes the CNN-based feature extractor (e.g. customized ResNet50 output 1024-channel), SCG module and K-layer GCNs (K=2 in this work), the fused (element-wise sum) output is projected back to 2D maps for final prediction.}
  \label{fig:scg_net}
\end{figure}

The SCG module can be easily incorporate into existing CNN and GCN architectures in order to exploit the advantages of both the CNN to learn feature detectors, while at the same time exploit the ability of GCNs to model long-range relations.
Fig.~\ref{fig:scg_net} shows our so-called SCG-Net, which combines SCG with CNNs and GCNs to address the semantic labeling task. Following our previous work \cite{liu_2019}, we utilize the first three bottleneck layers of a pretrained ResNet50 \cite{he2016deep} as the backbone CNN to learn the high-level representations. A 2-layer GCN (Equation~\ref{eq:gcn2}) is used in our model and we utilize ReLU activation and batch normalization only in the first layer of the $\operatorname{GCN}$.

\section{Experiments and results}
We train and evaluate our proposed methods on a publicly available benchmark dataset, namely the ISPRS 2D Vaihingen semantic labeling contest dataset. The Vaihingen dataset contains 33 tiles of varying size (on average approximately $2100 \times 2100$ pixels) with a ground resolution of 9cm, of which 17 are used as hold-out test images. We follow the training settings of our previous work \cite{liuqinghui2018} to train our model, and apply a dice loss function \cite{milletari2016v} and two regularization terms $\mathcal{L}_{kl}$ and $\mathcal{L}_{dl}$ as defined in the equations \ref{eq:klloss} and \ref{eq:dlloss}.
The overall cost function of our model is therefore defined as
\begin{equation}\label{eq:cost_function}
    \mathcal{L} \gets \mathcal{L}_{dice} + \mathcal{L}_{kl} + \mathcal{L}_{dl} \; .
\end{equation}

We train and validate the networks with 4000 randomly sampled patches of size $448\times 448$ as input and train it using minibatches of size $4$. The training data is sampled uniformly and randomly shuffled for each epoch.

\begin{figure*}[htpb!]
 \centering
  \includegraphics[width=0.93\textwidth]{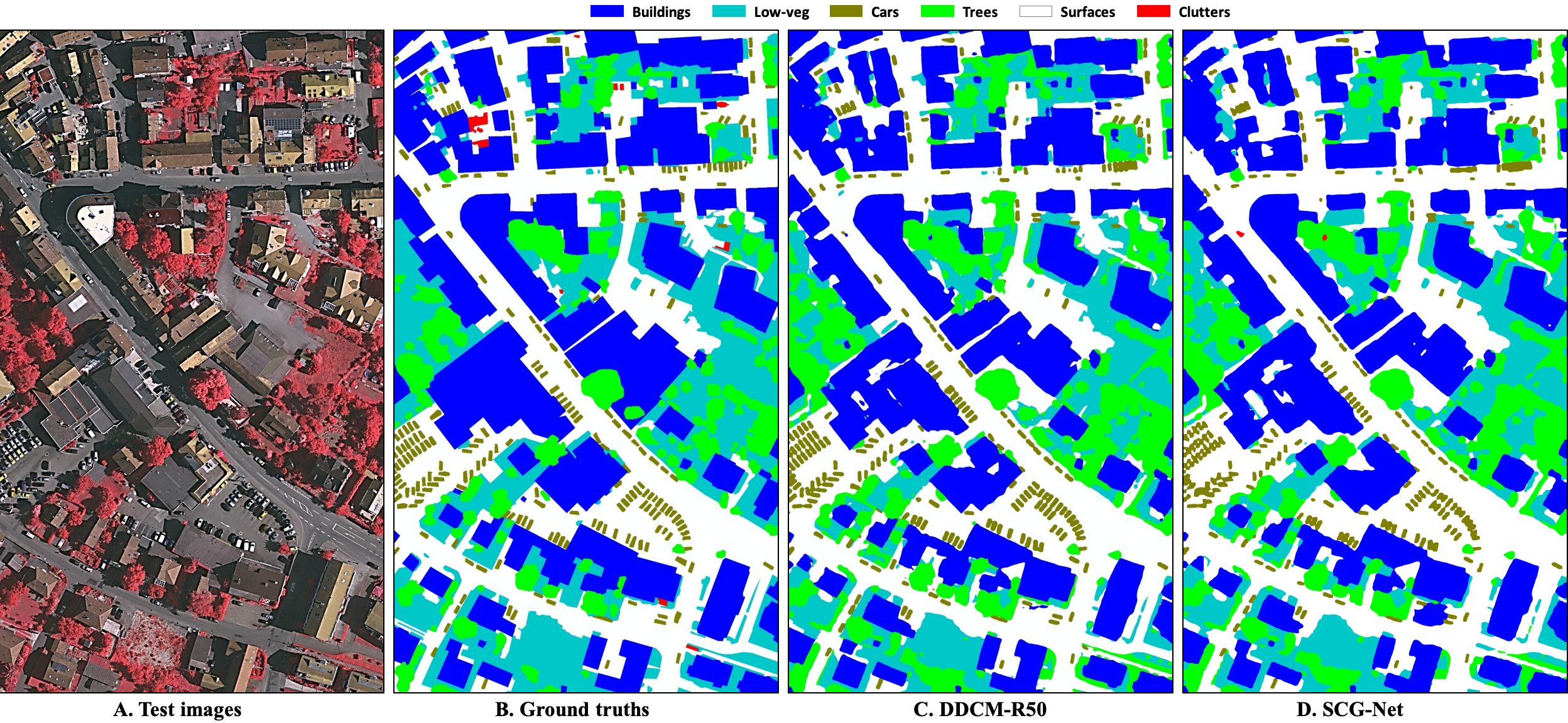} 
  \caption{Mapping results for test images of Vaihingen tile-27. From the left to right, the input images, the ground truths and the predictions of DDCM-R50, and our SCG-Net.}
  \label{fig:test}
\end{figure*}

\textbf{Results}:
We evaluated our trained model on the hold-out test sets (17 images) in order to fairly compare to other related published work on the same test sets. These results are shown in Table~\ref{tab:vaihingen_scores}. Our model obtained very competitive performance with 89.8\% F1-score which is around 1.1\% higher than GSN \cite{wang2017gated} and the same as the best performing model DDCM-R50 \cite{liu_2019}. However, the proposed model consists of fewer training parameters ($8.74$ million vs. $9.99$ million for the DDCM-R50 model) and has lower computational cost ($4.37$ Giga FLOPs vs. $4.86$ Giga FLOPs for the DDCM-R50 model), resulting in faster training performance. Fig.~\ref{fig:test} shows the qualitative comparisons of the land cover mapping results from our model and the ground truths on the test set. 




\begin{table}[hptb!]
\centering 
  \caption{Comparisons between our method with other published methods on the hold-out IRRG test images of ISPRS Vaihingen Dataset.} 
  
\resizebox{\columnwidth}{!}{
\begin{threeparttable}
\begin{tabular}{c|p{9mm}|p{9mm}p{10mm}p{13mm}p{9mm}p{9mm}|p{8mm}} \hline
    \textbf{Models} & $\textbf{OA}$ & \textbf{Surface} & \textbf{Building} & \textbf{Low-veg} & \textbf{Tree} & \textbf{Car}  & \textbf{mF1} \\  \hline \hline
    ONE\_7 \cite{audebert2016semantic}          & 0.898  & 0.910  & 0.945  & \textbf{0.844} & 0.899  & 0.778 & 0.875\\ 
    DLR\_9 \cite{MarmanisSWGDS16}               & 0.903  & 0.924  & 0.952  & 0.839 & 0.899  & 0.812  & 0.885 \\  
    GSN \cite{wang2017gated}                    & 0.903  & 0.922  & 0.951  & 0.837 & \textbf{0.899} & 0.824  & 0.887 \\ 
    DDCM-R50 \cite{liu_2019} & \textbf{0.904} & \textbf{0.927}& \textbf{0.953}    & 0.833   &0.894 & \textbf{0.883} & \textbf{0.898}  \\ \hdashline%
     $\textbf{SCG-Net}$ & \textbf{0.904}  & 0.924  & 0.948  & \textbf{0.839} & \textbf{0.897}  & 0.880  & \textbf{0.898} \\  \hline
\end{tabular}
  \end{threeparttable}
} 
\label{tab:vaihingen_scores}%

\end{table}

\section{Conclusions} \label{concl}
In this paper, we presented a self-constructing graph (SCG) architecture which makes use of learnable latent variables to construct the hidden graphs directly from 2D feature maps with no prior graphs available. The proposed SCG network can be easily adapted and incorporated into existing deep CNNs and GCNs architectures to address a wide range of different problems. On the Vahingen datasets, our SCG-Net model achieves competitive results, while making use of fewer parameters and being computationally more efficient. 

\bibliographystyle{IEEEbib}
\bibliography{my.bib}

\end{document}